\documentclass[runningheads]{llncs}

\usepackage{graphicx}

\usepackage{amsmath,amsfonts,amssymb}

\usepackage{hhline}
\usepackage{multirow}
\usepackage{booktabs}
\usepackage{wrapfig,lipsum}
\usepackage{hyperref}

\usepackage{color,xcolor}
\usepackage{xspace}

\usepackage{pifont}
\usepackage[super]{nth}

\usepackage{bm}

\usepackage[markup=changes]{changes}

\usepackage{siunitx}  %

\definecolor{citecolor}{RGB}{34,139,34}
\definecolor{mydarkblue}{rgb}{0,0.08,1}
\definecolor{mydarkgreen}{rgb}{0.02,0.6,0.02}
\definecolor{mydarkred}{rgb}{0.8,0.02,0.02}
\definecolor{mydarkorange}{rgb}{0.40,0.2,0.02}
\definecolor{mypurple}{RGB}{111,0,255}
\definecolor{myred}{rgb}{1.0,0.0,0.0}
\definecolor{mygold}{rgb}{0.75,0.6,0.12}
\definecolor{myblue}{rgb}{0,0.2,0.8}
\definecolor{mydarkgray}{rgb}{0.,0.2,0.2}

\newcommand{\figref}[1]{Fig.~\ref{#1}}
\newcommand{\tabref}[1]{Table~\ref{#1}}

\newcommand{\tabcaption}[1]{\caption{#1}}

\begin{document}
\title{Consistency Regularization Improves Placenta Segmentation in Fetal
EPI MRI Time Series}
\titlerunning{Consistency regularization improves placenta segmentation}

\author{Yingcheng Liu\inst{1} \and
Neerav Karani\inst{1} \and 
Neel Dey\inst{1} \and 
S. Mazdak Abulnaga\inst{1,4} \and
Junshen Xu\inst{2} \and
P. Ellen Grant\inst{3} \and
Esra Abaci Turk\inst{3} \and
Polina Golland\inst{1}
}

\authorrunning{Y. Liu et al.}

\institute{Computer Science and Artificial Intelligence Lab, Massachusetts Institute of Technology, Cambridge, MA, USA \and
Department of Electrical Engineering and Computer Science, Massachusetts Institute of Technology, Cambridge, MA, USA \and 
Fetal-Neonatal Neuroimaging and Developmental Science Center, Boston Children's Hospital, Harvard Medical School, Boston, MA, USA \and 
Martinos Center for Biomedical Imaging, Harvard Medical School, Boston, MA, USA
\\
\email{\{liuyc,nkarani,dey,abulnaga,junshen\}@mit.edu, \{ellen.grant,esra.abaciturk\}@childrens.harvard.edu, polina@csail.mit.edu}}

\maketitle              %

\begin{abstract}
The placenta plays a crucial role in fetal development. Automated 3D placenta segmentation from fetal EPI MRI holds promise for advancing prenatal care. This paper proposes an effective semi-supervised learning method for improving placenta segmentation in fetal EPI MRI time series. We employ consistency regularization loss that promotes consistency under spatial transformation of the same image and temporal consistency across nearby images in a time series. The experimental results show that the method improves the overall segmentation accuracy and provides better performance for outliers and hard samples. The evaluation also indicates that our method improves the temporal coherency of the prediction, which could lead to more accurate computation of temporal placental biomarkers. This work contributes to the study of the placenta and prenatal clinical decision-making. Code is available at \url{https://github.com/firstmover/cr-seg}

\end{abstract}

\section{Introduction}
In this paper, we propose an effective semi-supervised learning method that improves the segmentation of the placenta in fetal EPI MRI time series. The automatic 3D segmentation of the placenta facilitates better population studies~\cite{leon2018retrospective}, visualization of individual placentae for monitoring and assessment~\cite{abulnaga2019placenta}, and intervention planning. 

The promise of automated placenta segmentation in 3D MRI has been previously demonstrated~\cite{abulnaga2022automatic,alansary2016fast,melbourne2016placental}. Neural networks provide state-of-the-art performance and have emerged as the most popular paradigm~\cite{abulnaga2022automatic,alansary2016fast}. However, one shortcoming of deep learning methods is that they require a large number of annotated training examples. Manual segmentation of placenta is difficult in temporal 3D MRI due to inherently volumetric shape of the placenta and due to deformations caused by maternal breathing, contractions, and fetal motion. Furthermore, functional EPI images of the placenta have lower in-plane resolution than anatomical HASTE volumes. The contrast between the organ and the surrounding anatomy is worse, making it challenging to determine the placental boundary. These issues preclude annotation of large datasets, thereby hindering the development of accurate neural network models.

One possible solution to this problem is semi-supervised learning, which promises to yield performant neural network models while requiring only a partial labeling of the dataset. One line of work promotes consistency between labeled and unlabeled data. For example, encouraging consistency of the feature embedding distribution between labeled and unlabeled data regularizes the model's training either directly~\cite{baur2017semi} or via adversarial learning~\cite{kamnitsas2017unsupervised}. Another line of work uses unlabeled and labeled data sequentially in different training stages by first pre-training an image representation on a large unlabeled dataset and then fine-tuning the segmentation model with a small labeled dataset~\cite{taleb20203d,tang2022self}.

In this paper, we build on the third strategy of integrating semi-supervised learning into the training of the segmentation network. Our method is based on consistency regularization, a class of semi-supervised learning methods that promotes the invariance and equivariance of model predictions. The idea is to make the segmentation predictions consistent between two different augmented versions of the same image. We refer to this consistency as spatial consistency. Applying existing spatial consistency regularization methods~\cite{cui2019semi,li2020transformation,xia2010direct} to our data only emphasizes the consistency within the same image but ignores the relationships among different images in a time series. To encourage consistency across images, we require the deformations of the segmentations to be consistent with the deformations of the corresponding images. To estimate the deformation in images, we pretrain a registration model~\cite{balakrishnan2019voxelmorph}. This consistency regularization encourages the segmentation model to be equivariant under temporal deformations. Hence we refer to this consistency as temporal consistency. Temporal consistency was also studied in~\cite{ren2022local}. Their method only applies to MRI with 2 to 5 time points while we tackle more than one hundred time frames. 

To implement spatial and temporal consistency, we employ a two-branch Siamese architecture during training~\cite{chicco2021siamese}. Each branch in the Siamese network receives a different image as input (either spatially transformed versions of the same image or close frames in the time series) and is supervised to output predictions consistent with the other branch. 

We evaluate our method on a dataset of 3D EPI MRI placenta time series of 91 subjects. We show that our method provides modest but robust improvements on average, and significantly improves the quality of segmentation in the outlier cases. The results also suggest that our method is more sample-efficient than other semi-supervised learning approaches such as mean teacher~\cite{tarvainen2017mean} and self-training~\cite{yarowsky1995unsupervised}.

\section{Methods}

\begin{figure}[t]
\centering
\includegraphics[width=\columnwidth]{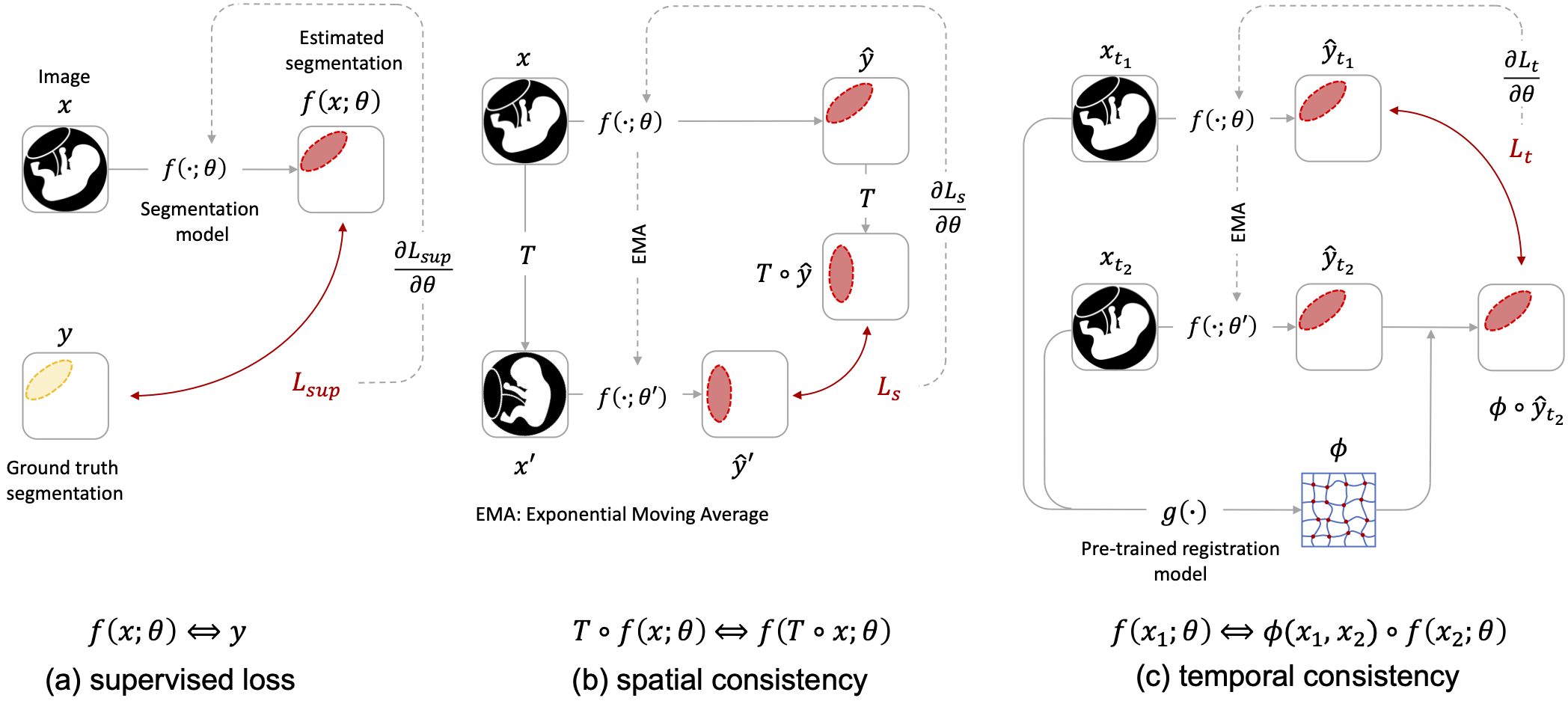}
\caption{\textbf{Overview of Consistency Regularization Training Scheme.} Our consistency regularization loss includes three parts: (a) supervised segmentation loss (b) spatial consistency regularization loss that promotes consistency within the same sample (c) temporal consistency regularization loss that encourages the predictions to be consistent between different frames in the same time series. When using the teacher-student training for (b) and (c), we use the exponential moving average (EMA) of the student model to update the teacher model.}
\label{fig:overview}
\end{figure}

In this section, we will first define spatial and temporal consistency regularization loss central to our method in Sec.~\ref{sec:cr_loss}, describe the Siamese training architecture and its teacher-student variant in Sec.~\ref{sec:siamese_nn}. Last, we provide implementation details in Sec.~\ref{sec:impl_detail}. The high-level overview of our training scheme is illustrated in \figref{fig:overview}.

\subsection{Consistency Regularization Loss} 
\label{sec:cr_loss}

Let $\mathcal{X}_l = \{(x_i,y_i), i = 1, 2, ..., n\}$ and $\mathcal{X}_u = \{x_i, i = 1, 2, ..., m\}$ be the labeled and unlabeled data, where $x_i \in \mathbb{R}^{H \times W \times D}$ is a 3D image and $y_i \in \{0, 1, 2, ..., C - 1\}^{H \times W \times D}$ is the ground truth segmentation label map with $C$ distinct classes ($C=2$ in our application). A neural network segmentation model predicts a segmentation map $\hat{y} = \sigma(f(x; \theta)) \in \mathbb{R}^{H \times W \times D}$ from an input image $x$, where $\sigma(\cdot)$ is a softmax function and $f(x; \theta)$ is the logits map of the neural network parameterized by $\theta$. 
Ground truth label maps are used to minimize supervised loss $\mathcal{L}_{\text{sup}}(\theta) = \mathbb{E}_{(x,y) \in \mathcal{X}_l} \mathcal{S}(f(x;\theta), y)$ for image segmentation loss function $\mathcal{S}(\cdot, \cdot)$

The central idea of consistency regularization is to explicitly encourage the segmentation model to be invariant or equivariant to a certain group of transformations. We apply this idea both to the same input image and across different images in the time series. 
First, we regularize the segmentation models to be equivariant under geometric transformations (e.g., rigid or elastic transformations) and invariant to intensity transformations (e.g., contrast, brightness, and Gaussian noise). During the training of the model, these transformations are drawn stochastically from a pre-defined transformation distribution $\mathcal{T}$ and applied to the image before it is provided as an input to the neural network. We aim to minimize the spatial consistency loss term 
\begin{equation}
\mathcal{L}_{\text{s}}(\theta) = 
{
\mathop{\mathbb{E}}_{T \sim \mathcal{T}, x \in \mathcal{X}_l \cup \mathcal{X}_u}
\ell \Big( 
T \circ f(x; \theta), f(T \circ x; \theta) 
\Big),
}
\end{equation}
where $\ell$ is a pixel-level consistency loss that measures the distance between two predicted logit maps.

For temporal consistency, we first randomly sample a frame $x$ in the time series and within the nearby $\Delta t$ frames, we randomly sample another frame $x'$. We estimate the inter-frame transformation $\phi$ using a pre-trained and frozen registration model $g(\cdot)$ on input images. We encourage the segmentation predictions for these two frames to be similar when aligned using $\phi$. This temporal consistency loss term can be written as
\begin{equation}
\mathcal{L}_{\text{t}}(\theta) = 
{
\mathop{\mathbb{E}}_{x \in \mathcal{X}_u}
\ell \Big( 
f(x; \theta), g(x, x') \circ f(x'; \theta) 
\Big),
}
\end{equation}
where $x$ and $x'$ are regarded as fixed and moving images, respectively, by the registration model. 
Since consistency losses do not rely on ground truth segmentation labels, all unlabeled data can be incorporated into the training process. 

Combining, we optimize parameters $\theta$ to minimize 
\begin{equation}
\begin{aligned}
\mathcal{L}(\theta) & = \mathcal{L}_{\text{sup}}(\theta) + \lambda_{1} \mathcal{L}_{\text{s}}(\theta; \mathcal{T}) + \lambda_{2} \mathcal{L}_{\text{t}}(\theta) \\ 
& = \mathop{\mathbb{E}}_{(x,y) \in \mathcal{X}_l}{\mathcal{S}(f(x;\theta), y)} + \lambda_{1} \mathop{\mathbb{E}}_{T \sim \mathcal{T}, x \in \mathcal{X}_l \cup \mathcal{X}_u} \ell \Big( T \circ f(x; \theta), f(T \circ x; \theta) \Big)\\
& + {\lambda_{2} \mathop{\mathbb{E}}_{x \in \mathcal{X}_u}\ell \Big( f(x; \theta), g(x, x') \circ f(x'; \theta) \Big)},
\end{aligned}
\end{equation}
where $\lambda_1, \lambda_2$ are regularization parameters.

\subsection{Siamese Neural Network} 
\label{sec:siamese_nn}

We use the U-Net architecture \cite{ronneberger2015unet} as our segmentation model. 
We employ the so-called Siamese architecture to minimize the consistency loss terms. Specifically, one copy of the network operates on the current input image $x$, while another copy takes augmented image $T \circ x$ for spatial consistency and next frame $x'$ for temporal consistency. Differences between the two segmentations are used to update the network weights. 
This training scheme has been widely applied to several problems such as object tracking \cite{bertinetto2016fully} and representation learning \cite{chen2021exploring}. 

Teacher-student variant of this Siamese design is known to be beneficial in many semi-supervised learning problems \cite{cui2019semi,li2020transformation}. Specifically, the ``teacher" branch of the model is decoupled from the gradient descent update process, and its weights are updated using the exponential moving average of the ``student" model, i.e., the weight of teacher model $\theta_{k}^{'}$  is updated using $\theta_{k}^{'} = \alpha \theta^{'}_{k-1} + (1 - \alpha) \theta_{k} $, where $\theta_{k}$ is the student model parameters in iteration $k$ of training, and $\alpha$ is a smoothing coefficient. Prior evidence shows that the teacher model has more stable predictions than the student model and provides more accurate supervision \cite{cui2019semi,li2020transformation}. Another practical consideration of this architecture is that the teacher-student network is more computationally efficient. Since the teacher model is updated through exponential moving average, it takes up minimal GPU memory in the training process and saves approximately half of the memory compared to methods that backpropagate through both branches of the Siamese architecture.

\subsection{Implementation Details} 
\label{sec:impl_detail}

In practice, we approximate the minimization of the objective above using batch stochastic gradient descent. At each step of the gradient descent, we sample a set of data $\mathcal{D} = \mathcal{D}_l \cup \mathcal{D}_{u_i}$, where $\mathcal{D}_l$ is uniformly sampled from labeled data $\mathcal{X}_l$ and $\mathcal{D}_{u_i}$ is uniformly sampled from unlabeled data $\mathcal{X}_u$. We make these datasets to have the same size (i.e., $|\mathcal{D}_l| = |\mathcal{D}_{u_i}|$). For each sample $x \in \mathcal{D}$, we independently draw a transformation $T$ from $\mathcal{T}$ and retrieve a nearby frame $x' \in \mathcal{D}_u$ with $\Delta t = 5$.

To construct image transformation $T$, we sequentially apply the following transformations: random flipping along all dimensions with $p=0.5$, random rotation along all dimensions with $p=0.5$, random translation along all dimensions with $p=0.1$ where displacements for each dimension are sampled from $\mathcal{U}(0, 5)$, random Gamma transform with $\gamma$ sampled from a uniform distribution $\mathcal{U}(0.5, 2)$, and random pixel-wise Gaussian noise with zero mean and $\sigma$ sampled from a uniform distribution $\mathcal{U}(0, 0.1)$. We encourage the logit maps to be equivariant under random flipping, rotation, and translation transform while invariant to the random gamma and Gaussian noise transforms. Specifically, the transformation of the logits map contains only the random flipping, rotation, and translation transform of the corresponding image transformation. 

 We modify $\lambda_1$ and $\lambda_2$ during the training following prior practice~\cite{cui2019semi,li2020transformation}. Specifically, we use $\lambda_i=\lambda_0 \exp\left(-5(1 - \frac {S}{L}) ^2 \right)$ when $S\leq L$ is the current training step and $L$ is the ramp-up length, $i=1,2$. When $S> L$, $\lambda_i$ is set to $\lambda_0$. We set $L$ to be half of the total training steps. The intuition behind this schedule is that the regularization is only necessary at the later stages of training where overfitting is more severe. We use three lambda values ($\lambda_i = 0.01, 0.001, 0.0001$) for $\lambda_0$. We report results for the best-performing model on the validation set.
 
 We used a sum of Dice loss and cross-entropy loss as our supervised segmentation loss and pixel-wise L2 distance between logits as our consistency loss. We use the Adam optimizer~\cite{kingma2014adam} with a learning rate of $10^{-4}$ and weight decay of $10^{-5}$. We trained the model for $5000$ epochs with batch size $16$. We used linear warmup for $10$ epochs and then a cosine annealing learning rate scheduler. All experiments are performed on a workstation with four NVIDIA 2080Ti GPUs. Each experiment takes about six hours to finish.

To achieve frame-to-frame registration for temporal consistency regularization, we pre-train registration model $g(\cdot)$ following the VoxelMorph approach~\cite{balakrishnan2019voxelmorph}. Specifically, our registration model is a 3D convolutional UNet architecture that takes as input the channel-wise concatenation of two frames (i.e., a moving frame and a fixed frame). The model predicts 3D deformation fields as output. During the training process, we optimize its parameters to maximize the local normalized cross-correlation~\cite{avants2008symmetric} between two randomly sampled frames of the same subject. To compute this cross-correlation, we used a local window size of $(5,5,5)$. In addition, to encourage a smooth deformation field, we use an L2 regularizer on the spatial gradient of the deformation field. We re-train the registration model for each cross-validation experiment. The unlabeled training data set $\mathcal{D}_u$ only includes subjects that are also in the labeled data set $\mathcal{D}_l$.

\section{Experiments}

In this section, we evaluate the proposed consistency regularization method on a set of research fetal EPI MRI time series. 

\subsection{Data}

Our dataset consists of EPI MRI scans of $91$ subjects, of which $78$ were singleton pregnancies (gestational age (GA) of $23$wk$5$d -- $37$wk$6$d), %
and $13$ were monochorionic-diamniotic (Mo-Di) twins (GA at MRI scan of $27$wk$5$d -- $34$wk$5$d). The cohort includes $63$ healthy, $16$ fetal growth restriction (FGR), %
and $12$ high BMI (BMI $>30$) pregnancies. 

MRI scans were acquired on a $3$T Siemens Skyra scanner (GRE-EPI, interleaved with $3$mm isotropic voxels, TR $=5.8$--$8$s, TE $=32-47$ ms, FA = $90^{\circ})$. We split the acquired interleaved volumes into two separate volumes with spacing $3\times 3 \times 6$mm, then linearly interpolated to  $3 \times 3 \times 3$mm. 

The median length of the time series was 216 frames with an interquartile range (IQR) of 125. The placenta was manually segmented by a trained rater. For each MRI time series, between $1$ to $6$ EPI volumes were manually segmented, yielding a total of $176$ ground truth labelmaps. 

\subsection{Baseline Methods}

To understand the effect of different components of the consistency loss, we train the model with four different settings: basic model trained on labeled data using only the supervised loss $L_{\text{sup}}$ ({\it Basic}), model trained using segmentation loss with spatial transformation consistency regularization $L_{\text{sup}} + \lambda L_\text{s}$ on labeled data only ({\it S}), model trained using segmentation loss with spatial transformation consistency regularization  $L_{\text{sup}} + \lambda L_\text{s}$ on both labeled and unlabeled data ({\it S+}), and model trained using segmentation loss with spatial and temporal consistency regularization  $L_{\text{sup}} + \lambda_1 L_\text{s} + \lambda_2 L_\text{t}$ on both labeled and unlabeled data ({\it S+T}). 

We also compare our consistency regularization method with two representative semi-supervised learning methods: mean teacher \cite{tarvainen2017mean} and self-training \cite{yarowsky1995unsupervised}. In mean teacher, we implement the model using a similar Siamese student teacher architecture where the student model is supervised to be consistent with the teacher model. However, we do not ask the model to be invariant or equivariant to image transformations. In self-training, we first train the segmentation model using labeled dataset and impute segmentations of the unlabeled images using the prediction of this model. We then train a second model using this augmented dataset. We compare these three methods using both full and partial datasets. To create partial dataset, we randomly sub-sample training and validation dataset to contain $5,10,20,40,60$ subjects while keeping the test set unchanged.

\begin{figure}[t]
\centering
\includegraphics[width=\columnwidth]{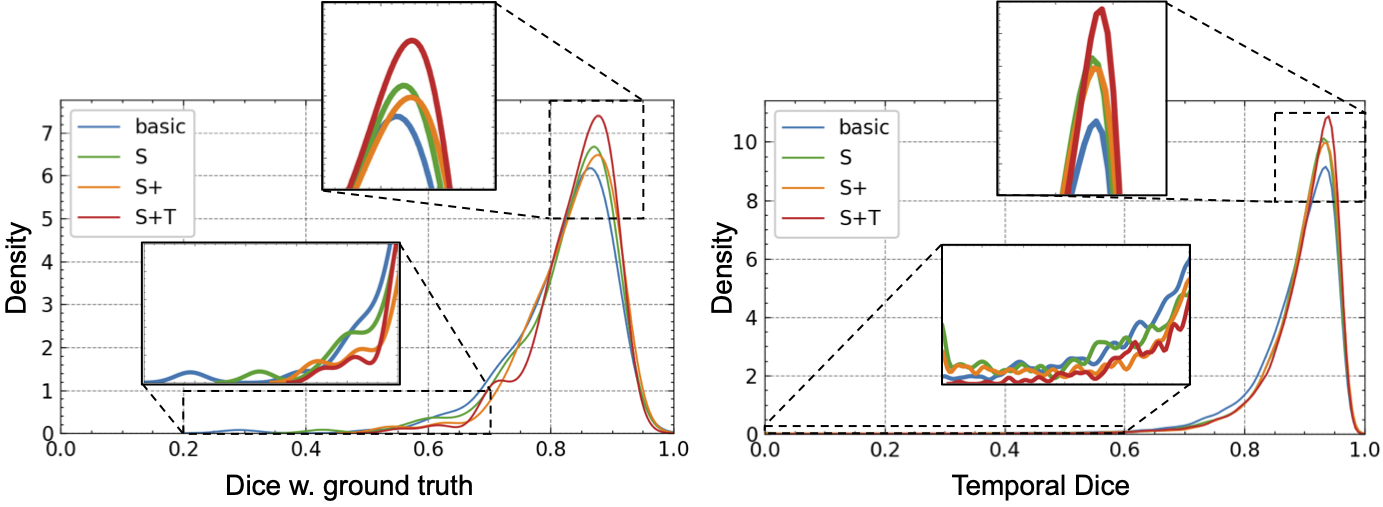}
    \caption{\textbf{Distribution of Volume Overlap for Ablation Experiments.} Left: Kernel density estimation of Dice w. ground truth for {\it Basic}, {\it S}, {\it S+}, and {\it S+T} settings. Right: Kernel density estimation of Temporal Dice in the same four training settings. For both metrics, we observe a tighter mode and a narrower long-tail distribution after applying our consistency regularization loss.}
\label{fig:histogram}
\end{figure}

\subsection{Evaluation}

To evaluate segmentation accuracy, we measure the volume overlap between the predicted and ground truth segmentation of placenta using the Dice similarity coefficient. We refer to this metric as Dice w. ground truth.
In addition, we assess the consistency of our predictions across time. We apply our model to all volumes in the time series and measure the consistency of consecutive predictions using Dice similarity. We refer to this metric as temporal Dice. While we do not expect perfect alignment in the presence of motion, relatively small motion of the placenta should yield a rough agreement. Thus, consecutive frames with low volume overlap are suggestive of segmentation errors. This provides a way to evaluate the segmentation quality without the excessive overhead of manual segmentation of the whole series. 
We also divide subjects into LOW and HIGH groups and report the Dice w. ground truth and temporal Dice per group. The HIGH group is defined as subjects whose performances are above threshold $\alpha$ in {\it Basic} training; LOW group is defined as those below. LOW group represents outliers and hard samples. We used $\alpha=0.8$ for Dice w. ground truth and $\alpha=0.7$ for temporal Dice. 

We use cross-validation to select the best hyperparameters. We randomly divide the subjects into five folds. We then train the model on four folds and evaluate the model on the remaining fold. We repeat this process five times and report the mean and standard deviation of the performance across all subjects. We used the same hyperparameters for all folds. The unlabeled training data is collected only from the labeled training data to ensure no subject overlap exists between training and testing sets. 

\begin{figure}[t]
\centering
\includegraphics[width=\columnwidth]{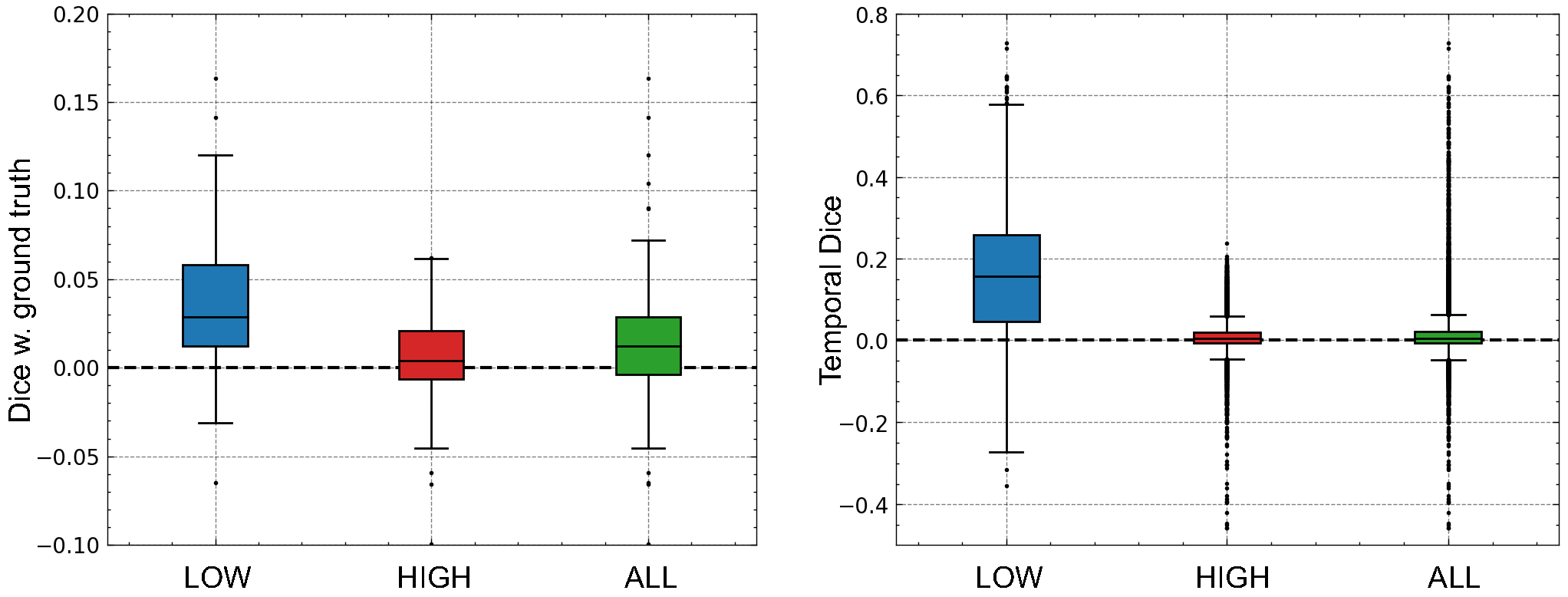}
\caption{\textbf{Boxplots of Dice w. ground truth and temporal Dice among three groups.} For S+T vs. Basic setting, we show boxplots of Dice w. ground truth and temporal Dice among LOW, HIGH, and ALL groups. We observe larger improvements in both metrics in LOW group.}
\label{fig:boxplot_low_high_all}
\end{figure}

\begin{table*}[t]
\centering
\setlength{\tabcolsep}{1.5pt}
\begin{tabular}{l|ccc|ccc}
    \toprule

    & \multicolumn{3}{c}{Dice w. ground truth} & \multicolumn{3}{c}{Temporal Dice} \\
    & LOW & HIGH & ALL & LOW & HIGH & ALL \\

    \midrule

    S vs. Basic & $0.015$\scriptsize{$\pm0.06$} & $0.0060$\scriptsize{$\pm0.02$} & $0.008$\scriptsize{$\pm 0.03$} & $0.060$\scriptsize{$\pm 0.19$} & $0.003$\scriptsize{$\pm 0.03$} & $0.003$\scriptsize{$\pm 0.03$} \\
   
    \midrule

    S+ vs. Basic & $0.022$\scriptsize{$\pm 0.06$} & $\boldsymbol{0.0072}$\scriptsize{$\pm 0.02$} & $0.010$\scriptsize{$\pm 0.03$} & $0.050$\scriptsize{$\pm 0.19$} & $0.003$\scriptsize{$\pm 0.03$} & $0.004$\scriptsize{$\pm 0.03$} \\
     
    \midrule

    S+T vs. Basic & $\boldsymbol{0.028}$\scriptsize{$ \pm 0.05$} & $0.0059$\scriptsize{$\pm 0.03$} & $\boldsymbol{0.012}$\scriptsize{$\pm 0.03$} & $\boldsymbol{0.158}$\scriptsize{$\pm 0.21$} & $\boldsymbol{0.005}$\scriptsize{$\pm 0.03$} & $\boldsymbol{0.005}$\scriptsize{$\pm 0.03$}\\

    \bottomrule

\end{tabular}
\vspace{0.1cm}
\tabcaption{\textbf{Sample-wise Increase of Dice Score among Different Groups.} We report the median (IQR) increase in Dice score relative to the {\it Basic} setting. We observe a moderate increase of Dice scores for ALL subjects, and large increase of Dice score for the LOW group. }
\label{tab:group_increase_of_performance}
\end{table*}

\subsection{Results}

\begin{figure}[t]
\centering
\includegraphics[width=\columnwidth]{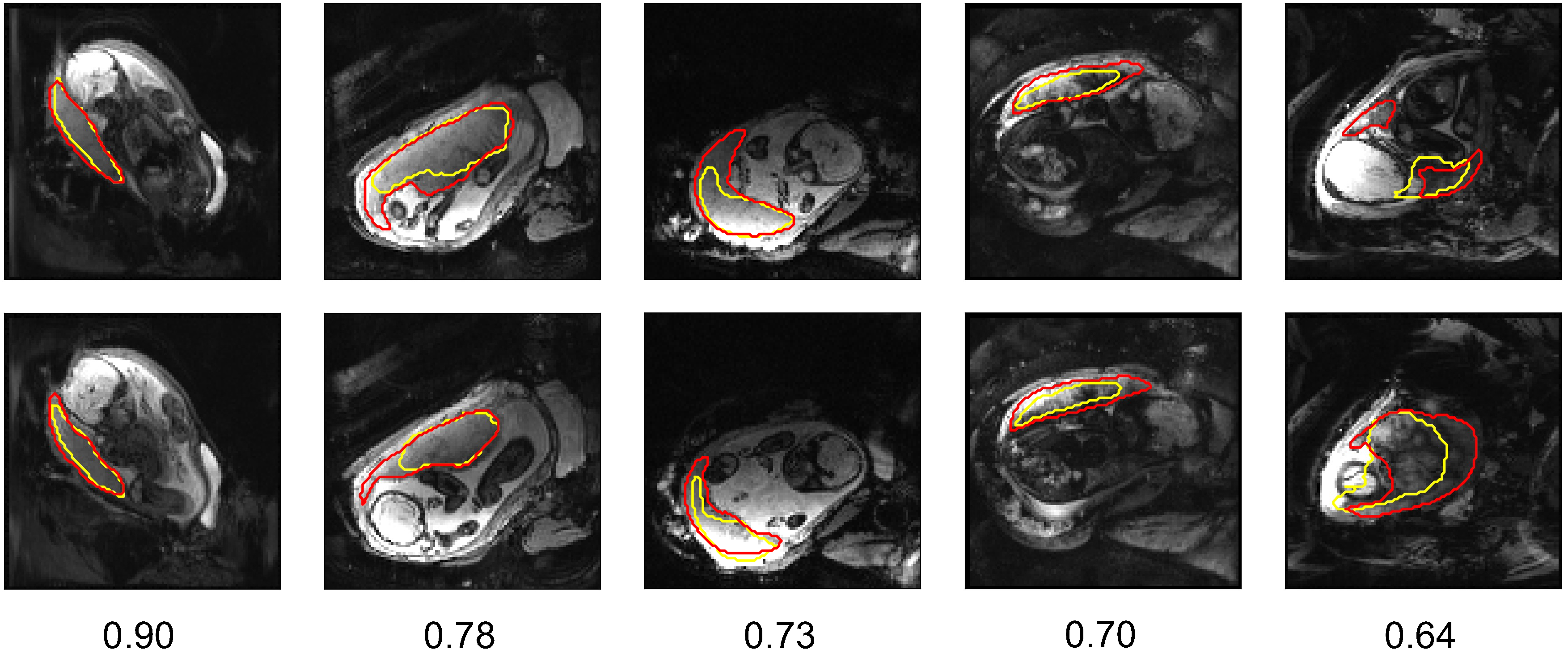}
\caption{\textbf{Example Predictions of Five Subjects across Different Test Performance.} Yellow indicates ground truth segmentation and red indicates predictions. Each column represents the same subject. Top and bottom slices are 18mm apart. On the bottom is the Dice score of the entire image.}
\label{fig:example_predictions}
\end{figure}

\begin{table*}[!ht]
\setlength{\tabcolsep}{3pt}
\centering
\begin{tabular}{c|ccccccc}
    \toprule

    Method & all & 60 & 40 & 20 & 10 & 5 \\

    \midrule

    mean teacher & $0.80$\scriptsize{$\pm0.09$} & $0.79$\scriptsize{$\pm0.07$} & $0.78$\scriptsize{$\pm0.08$} & $0.72$\scriptsize{$\pm0.13$} & $0.55$\scriptsize{$\pm0.21$} & $0.40$\scriptsize{$\pm0.19$} & \\
   
    \midrule

    self-training & $0.82$\scriptsize{$\pm0.09$} & $0.80$\scriptsize{$\pm0.07$} & $0.78$\scriptsize{$\pm0.08$} & $0.70$\scriptsize{$\pm0.10$} & $\boldsymbol{0.56}$\scriptsize{$\pm0.21$} & $0.35$\scriptsize{$\pm0.25$} & \\

    \midrule
    \midrule

    ours & $\boldsymbol{0.84}$\scriptsize{$\pm0.06$} & $\boldsymbol{0.83}$\scriptsize{$\pm0.09$} & $\boldsymbol{0.79}$\scriptsize{$\pm0.09$} & $\boldsymbol{0.76}$\scriptsize{$\pm0.16$} & $0.54$\scriptsize{$\pm0.18$} & $\boldsymbol{0.43}$\scriptsize{$\pm0.26$} &  \\
     
    \bottomrule

\end{tabular}
\vspace{0.1cm}
\tabcaption{\textbf{Segmentation Accuracy of SSL Methods in Different Levels of Data Availability.} We compare our consistency regularization with mean teacher~\cite{tarvainen2017mean} and self-training~\cite{yarowsky1995unsupervised} in both full and partial dataset settings. Our consistency regularization outperform both methods in most cases.}
\label{tab:compare_ssl}
\end{table*}

\figref{fig:histogram} presents the Dice distribution of models trained in four different settings. After applying consistency regularization, we observe a tighter mode and a narrower long-tail distribution. \tabref{tab:group_increase_of_performance} reports sample-wise improvement in volume overlap relative to the {\it Basic} for LOW, HIGH, and ALL groups. Our method provides moderate improvement for both groups and metrics (volume overlaps with ground truth and agreement in consecutive frames). The improvement is more pronounced for LOW group. In \figref{fig:boxplot_low_high_all}, we visualize the Dice improvement of {\it S+T} relative to {\it Basic} among three groups. The improvement in the LOW group is significantly higher than that in the HIGH and ALL groups. This indicates that our method improves the robustness of outlier and hard samples. We also visualize several predictions from our model in \figref{fig:example_predictions}. 

\tabref{tab:compare_ssl} compares our approach to the mean teacher~\cite{tarvainen2017mean} and self-training~\cite{yarowsky1995unsupervised} in different levels of data availability. Our method outperforms both methods in most of the settings.

\section{Limitations and future work}

There are several limitations to this study. First, our unlabeled samples are highly correlated with the labeled samples. This is in contrast with the typical semi-supervised learning settings where unlabeled samples are drawn independently from the same distribution as the labeled samples. This partly explains our experimental results where using unlabeled samples introduced relatively small improvements. Future studies should include images from new subjects. Second, several novel loss functions to improve segmentation in fetal MRI have been proposed recently~\cite{xu2020semi,abulnaga2022automatic}. We did not study how our consistency regularization method interacts with these methods. 

\section{Conclusions}

This work demonstrated a novel consistency regularization method that improves the performance and robustness of placenta segmentation in EPI MRI time series. The consistency regularization loss explicitly encourages the model to be consistent between transformations of the same image and between nearby frames in the time series. The evaluation shows that the method improves the overall segmentation accuracy and provides better performance for outlier and hard samples. Our method also improves the temporal coherency of the prediction. This could support the study of placenta and ultimately lead to more accurate placental biomarkers. 

\section*{Acknowledgements}
This research is supported by NIH NIBIB NAC P41EB015902, NIH NICHD R01HD100009, and NIH NIBIB 5R01EB032708, and the Swiss National Science Foundation project P500PT-206955.
\bibliographystyle{splncs04}
\bibliography{main.bib}

\newpage
\appendix

\section{Additional Results}

\begin{figure}[!htb]
\centering
\vspace{-1em}
\includegraphics[width=\columnwidth]{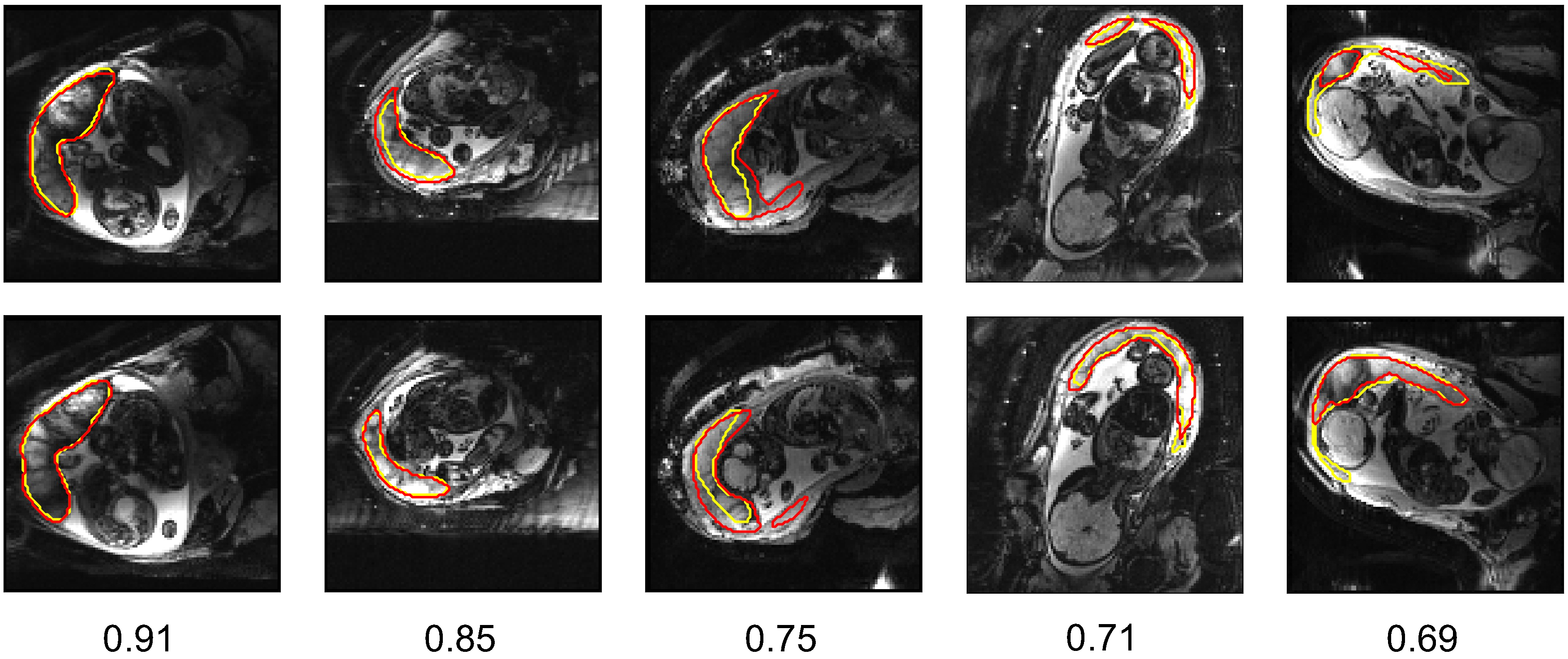}
\vspace{-1em}
\caption{\textbf{Additional Example Predictions.} Yellow indicates ground truth segmentation and red indicates predictions. Each column represents the same subject. Top and bottom slices are 18mm apart. On the bottom is the Dice score for the entire volume.}
\label{fig:example_prediction_supp}
\end{figure}

\begin{figure}[!htb]
\centering
\vspace{-2em}
\includegraphics[width=0.8\columnwidth]{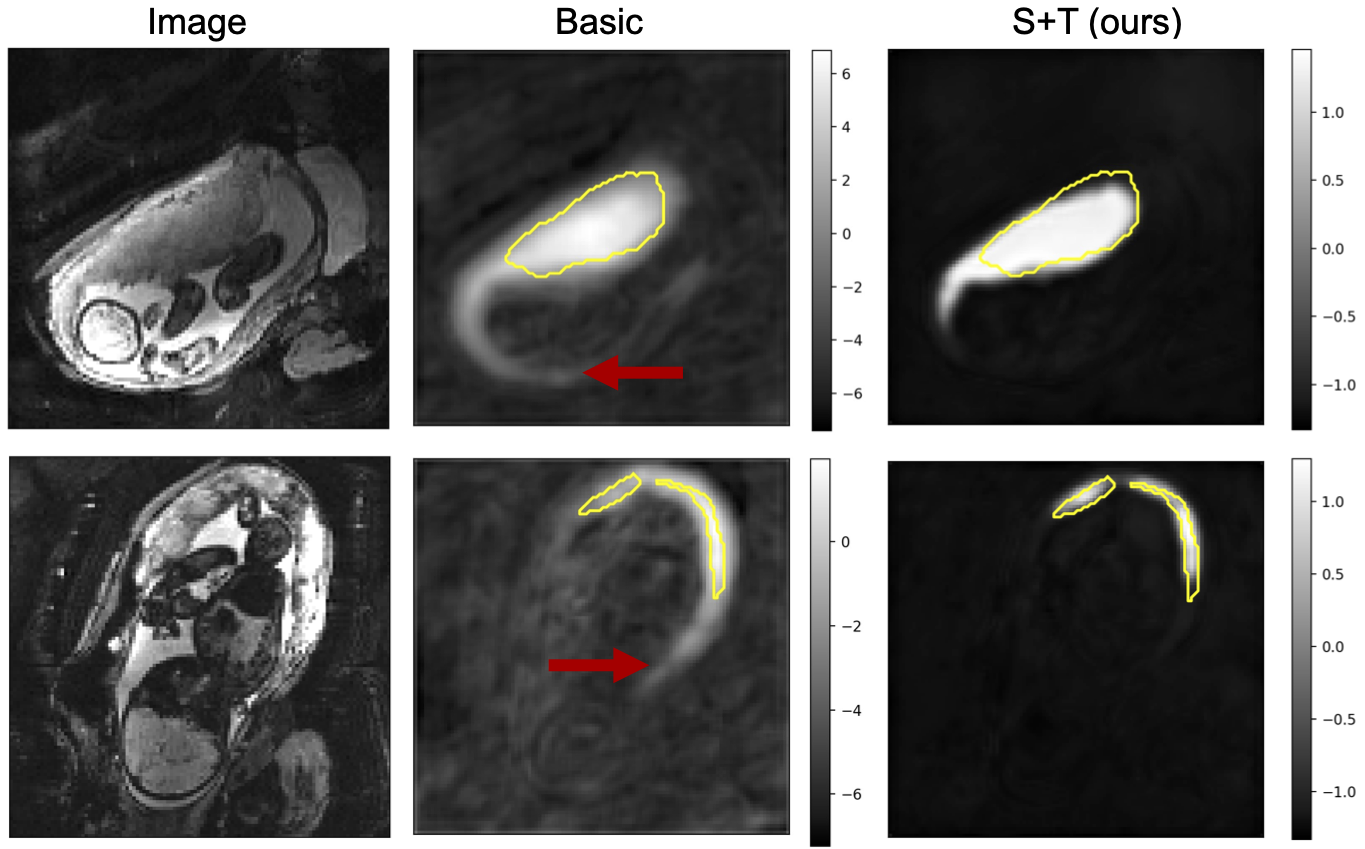}
\vspace{-0.5em}
\caption{\textbf{Visualization of Logits Map.} We compare the logits map of Basic and S+T models on two example images. White indicates high logit values. The yellow contour indicates ground truth segmentation. Two observations can be made: (1) Basic model produces higher logit values overall than the proposed S+T model. This corresponds to higher confidence predictions in the Basic model as compared to the S+T model. This observation is in line with recent work~\cite{karani2023boundary} which shows how consistency regularization based methods produce predictions with reduced confidence. (2) As indicated by red arrows, the high logit values of the Basic model are also present in non-placental area such as uterus wall while S+T’s predictions are more concentrated around the placental area. }
\label{fig:visualization_logits_map}
\end{figure}

\end{document}